\pdfoutput=1

\documentclass[11pt]{article}

\usepackage[]{naacl2021}

\usepackage{times}
\usepackage{latexsym}

\usepackage[T1]{fontenc}

\usepackage[utf8]{inputenc}

\usepackage{microtype}
\usepackage{colortbl}
\usepackage{graphicx}
\definecolor{dGray}{gray}{.6}
\definecolor{mGray}{gray}{.9}
\definecolor{lGray}{gray}{.94}
\usepackage{subfigure}
\usepackage{soul}
\soulregister\cite7
\soulregister\shortcite7 
\soulregister\ref7 

\newcommand*\multiplefootnoteseparator{%
  \textsuperscript{\multfootsep}\nobreak
}
\newcommand*\multfootsep{,}

\title{Modeling the Severity of Complaints in Social Media}

\author{Mali Jin \quad Nikolaos Aletras \\
  Department of Computer Science\\
  University of Sheffield, UK\\
  \texttt{\{mjin6, n.aletras\}@sheffield.ac.uk }
  }

\begin{document}
\maketitle
\begin{abstract}


\end{abstract}
The speech act of complaining is used by humans to communicate a negative mismatch between reality and expectations as a reaction to an unfavorable situation. Linguistic theory of pragmatics categorizes complaints into various severity levels based on the face-threat that the complainer is willing to undertake. This is particularly useful for understanding the intent of complainers and how humans develop suitable apology strategies. In this paper, we study the severity level of complaints for the first time in computational linguistics. To facilitate this, we enrich a publicly available data set of complaints with four severity categories and train different transformer-based networks combined with linguistic information achieving 55.7 macro F1. We also jointly model binary complaint classification and complaint severity in a multi-task setting achieving new state-of-the-art results on binary complaint detection reaching up to 88.2 macro F1. Finally, we present a qualitative analysis of the behavior of our models in predicting complaint severity levels.\footnote{Data is available here: \url{https://archive.org/details/complaint_severity_data}}\multiplefootnoteseparator\footnote{Code is available here: \url{https://github.com/mali726/Complaint-Severity}}



\section{Introduction}

Complaining is a speech act that usually conveys negative emotions triggered by a discrepancy between reality and expectations towards an entity or event \citep{olshtain1985complaints}. Complaints play an important role in human communication for expressing dissatisfaction. Based on complainers' personalities and specific situations, expression of complaints vary from person to person \citep{vasquez2011complaints}. 


In pragmatics, complaints have been classified into various levels of severity according to their emotional intensity, the amount of face-threat that the complainer is willing to undertake and their purpose \citep{olshtain1985complaints,trosborg2011interlanguage,kakolakigender}. Complaining purposes might include the expression of dissatisfaction, to find solutions (e.g. ask for reparations) or both. Furthermore, a complaint can be categorized as implicit (i.e. without mentioning who is responsible) or explicit (i.e. accusing someone for doing something). 

\renewcommand{\arraystretch}{1.2}
\begin{table*}[!t]
\small
\center
\resizebox{\textwidth}{!}{
\begin{tabular}{|l|p{12cm}|}
\hline
\rowcolor{dGray} \bf Label & \bf Example\\
\hline
No Explicit Reproach & \emph{Are you following me? I seem unable to send you a dm.}\\
\hline
\rowcolor{lGray} Disapproval & \emph{So far , the mac graphics drivers have been another disappointing update (for both my quadro 4000 \& gtx - 285),}\\
\hline
Accusation & \emph{Can u stop adding the UK keyboard layout to my Italian keyboard at every update? ktnxby}\\
\hline
\rowcolor{lGray} Blame & \emph{Thanks to $<$USER$>$ 's incompetence i now can't work till October 4th, when the ati card arrives.}\\
\hline
\end{tabular}}
\caption{Examples of complaint severity levels \citep{trosborg2011interlanguage}.}
\label{t:labelExample}
\end{table*}

Recent work on modeling complaints in natural language processing (NLP) has focused on distinguishing complaints from non-complaints in social media \citep{preotiuc2019automatically,jin2020complaint}, however there is no previous study into more fine-grained complaint categories. Table~\ref{t:labelExample} shows examples of social media posts expressing complaints grouped into four severity classes according to \citet{trosborg2011interlanguage}: (a) no explicit reproach; (b) disapproval; (c) accusation; and (d) blame. 

Identifying and analyzing the severity of complaints is important for: (a) improving customer service by recognizing the level of dissatisfaction and understanding complainers' needs~\citep{van2012online}; (b) linguists to study the speech act of complaints in different levels of granularity on large scale~\citep{tatsuki2000if}; and (c) developing downstream NLP applications such as automatic complaint response generation~\citep{Xu2017new} or voting stance prediction~\cite{tsakalidis2018nowcasting}.

In this paper, we present a systematic study on analyzing complaint categories with computational methods for the first time in computational linguistics. Our main contributions are as follows:
\begin{itemize}
    \item Grounded in linguistic theory of pragmatics \citep{trosborg2011interlanguage}, we enrich a publicly available data set~\citep{preotiuc2019automatically} with four complaint severity levels;
    \item We create a new classification task for identifying different severity levels of complaints;
    \item We evaluate transformer-based classification models \citep{vaswani2017attention} combined with linguistic information on (a) complaint severity level classification; and (b) binary complaint detection in a multi-task setting achieving new state-of-the-art results.
\end{itemize}

\section{Related Work}

 \subsection{Linguistic Categories of Complaints} 
 
 Previous work in linguistic theory of pragmatics has classified complaints into different levels based on their severity and directness. \citet{olshtain1985complaints} classified complaints into five distinct categories: (a) below the level of reproach; (b) expression of annoyance or disapproval; (c) explicit complaint; (d) accusation and (e) warning, immediate threat. More recently, \citet{trosborg2011interlanguage} proposed four major severity levels: (a) no explicit reproach; (b) disapproval; (c) accusation and (d) blame. Finally, \citet{kakolakigender} classified complaints into levels of directness: (a) very direct; (b) somewhat direct and (c) indirect. Direct complaints (i.e. very direct and somewhat direct) include obvious breaches of expectations. On the other hand, indirect complaints do not explicitly mention or can imply the breach of expectations. Moreover, the difference between very direct and somewhat direct is that the former highlights the responsibility of the complaint receiver while the latter does not. 

\subsection{Complaint Analysis}

Most of the existing studies on complaint classification in NLP have explored different approaches to the complaint identification task (identifing  complaints from non-complaints) in various domains, starting with feature-based machine learning models \citep{coussement2008improving, preotiuc2019automatically} and deep learning methods \citep{jin2020complaint}. \citet{coussement2008improving} used boosting ensemble models with linguistic style features to identify complaints in company emails. \citet{preotiuc2019automatically} applied logistic regression with a broad range of features to detect complaints in Twitter. More recent, \citet{jin2020complaint} explored a battery of transformer-based architectures combined with sentiment and topic information for complaint identification in social media. Also, previous work has classified complaints into detailed topical categories \citep{forster2017cognitive, merson2017text} or responsible departments \citep{laksana2014indonesian, gunawan2018building, tjandra2015determining}. Furthermore, other complaint related categorizations are based on product hazards and risks \citep{bhat2017identifying}, service failure \citep{jin2013service} and escalation likelihood \citep{yang2019detecting}.

\subsection{Emotion Detection}


Most related to complaint severity is emotion detection and its intensity which have been extensively studied in NLP~\citep{danisman2008feeler,volkova2016inferring,zhang2018ijcai}. More recently, \citet{alejo2020cross} explored cross-lingual transfer approaches to predict emotion intensity in Twitter. Similarly, \citet{akhtar2020intense} evaluated a series of feature-based machine learning models for both emotion and sentiment intensity prediction in social and news media.

\section{Task \& Data}

We define complaint severity prediction as a multi-class classification task. Given a text snippet $T$, defined as a sequence of tokens $T=\{t_1, ..., t_n\}$, the aim is to classify $T$ as one of the four predefined severity labels.

We use an existing complaints data set developed by \citet{preotiuc2019automatically}, which consists of 1,235 complaints (35.8\%) and 2,214 non-complaints (64.2\%) in English. We opted using this data set because it is publicly available with annotated complaints collected from Twitter in 9 general domains (i.e. Food, Apparel, Retail, Cars, Service, Software, Transport, Electronics and Other). 

\subsection{Complaint Severity Categories}

For complaint severity annotation, we adopt the four categories defined by \citet{trosborg2011interlanguage} because it is considered as the `standard' in pragmatics literature (see examples in Table~\ref{t:labelExample}): 

\begin{itemize}
    \item {\bf No explicit reproach:} there is no explicit mention of the cause and the complaint is not offensive;
    \item {\bf Disapproval:} express explicit negative emotions such as dissatisfaction, annoyance, dislike and disapproval; 
    \item {\bf Accusation:}  asserts that someone did something reprehensible; 
    \item {\bf Blame:} assumes the complainee is responsible for the undesirable result.
\end{itemize}

Note that the severity levels categorize complaints by type instead of intensity. Classes are disjoint according to \citet{trosborg2011interlanguage}. More specifically, `No explicit reproach' is a suggestive strategy, where the complainee is usually not mentioned in the statement. `Disapproval' expresses negative sentiment or unsatisfying state only. The statement may imply the complainer holds the complainee responsible but avoid mentioning it, which is the key component of identifying `Disapproval' and `Accusation'/`Blame'. The main difference between `Accusation' and `Blame' is in the latter one the complainer presupposes the complainee is guilty of the offense.  



\subsection{Complaint Severity Annotation}

Following the definitions above, each tweet was labeled by three annotators independently. In case of ties, the final decision was made by the authors through consensus. We recruited 35 native English speaking annotators from the volunteers list of our institution.\footnote{We have received approval from the Ethics Committee of our institution.}\multiplefootnoteseparator\footnote{Annotators are provided with an introduction of the task including definitions and examples of each category.} The inter-annotator agreement between the three original annotations for each tweet is $k=0.64$ Fleiss' Kappa\footnote{We randomize the order of three annotations for each tweet three times and compute the average Fleiss’ Kappa.} \citep{fleiss1971measuring} which belongs to substantial agreement \citep{artstein2008inter}. 

Table~\ref{t:labelCount} shows the distribution of tweets across classes: 435 tweets belong to `No Explicit Reproach' (35.2\%), 378 belong to `Disapproval' (30.6\%), 225 belong to `Accusation' (18.2\%); and 197 belong to `Blame' (16.0\%). The class distributions over 5 domains (Car, Retail, Service, Software, Transport) are similar to the overall distribution while 4 domains (Food, Apparel, Electronics, Other) differ from Table \ref{t:labelCount}. In domains with different distribution, differences appear especially in `No Explicit Reproach' and `Accusation', which might result from domain specific complaint requests.

\subsection{Text Processing}
Text is processed by lower-casing, and replacing all mentions of usernames and URLs with placeholder tokens. A Twitter-aware tokenizer, DLATK \citep{schwartz2017dlatk}, is used for text tokenization to handle emoticons and hashtags in social media text.



\renewcommand{\arraystretch}{1.2}
\begin{table}[!t]
\small
\center
\begin{tabular}{|l|c|c|}
\hline
\rowcolor{dGray} \bf Labels & \bf Amount & \bf Percentage\\
\hline
No Explicit Reproach & 435 & 35.2\\
\hline
Disapproval & 378 & 30.6\\
\hline
Accusation & 225 & 18.2\\
\hline
Blame & 197 & 16.0\\
\hline
\rowcolor{lGray} Total & 1235 & 100\\
\hline
\end{tabular}
\caption{Number of tweets in different complaint severity levels and class distribution.}
\label{t:labelCount}
\end{table}

\section{Predictive Models}

Since severity complaint prediction is a new task, we first evaluate the majority class as well as three strong baselines: (1) logistic regression with bag-of-words; (2) a bidirectional recurrent neural network trained from scratch; and (3) finetuning a pretrained transformer-based model. Furthermore, we combine linguistic information (i.e. emotion and topic information) to a transformer-based model similar to the method proposed by \citet{jin2020complaint} in the context of binary complaint classification. 

\subsection{Baselines}

\paragraph{Majority Class} We use Majority Class as the first baseline, where we calculate scores by labeling all the tweets with the majority class. 

\paragraph{LR-BOW} We use a linear baseline, Logistic Regression with standard bag-of-words (LR-BOW) and L2 regularization.

\paragraph{BiGRU-Att} We also use a neural baseline trained from scratch; a bidirectional Gated Recurrent Unit (GRU) network \citep{cho2014learning} with a self-attention mechanism (BiGRU-Att; \citep{tian2018attention}). Given a Twitter post $T$, a token $t_i$ is mapped to a GloVe embedding \citep{pennington2014glove}. We then apply dropout to the output of GloVe embedding layer and pass it to a bidirectional GRU with self-attention layer. Finally, the contextualized token representations are passed to an output layer using a softmax activation function for multi-class classification.

\paragraph{RoBERTa}
Bidirectional Encoder Representations from Transformers (BERT) \citep{devlin2018bert} is a pre-trained language model based on the Transformer architecture \citep{vaswani2017attention}. It makes use of multiple multi-head attention layers to learn context information from both the left and the right side of tokens. It is trained on masked language modeling by randomly masking some of the tokens from the input aiming to predict them based on the context only. RoBERTa \citep{liu2019roberta} is an extension of BERT trained on more data with different hyperparameters and has achieved better performance in social media analysis tasks~\cite{maronikolakis2020analyzing}. We fine-tune RoBERTa\footnote{We 
only report the results of RoBERTa because it achieves better performance compared to BERT over all evaluation methods in our experiments.} on complaint severity classification by adding an output dense layer with a softmax activation function.

\subsection{M-RoBERTa with Linguistic Information}

Multimodal-BERT (M-BERT) \citep{rahman2019m} injects multimodal information such as image and speech into the text representations of BERT. It combines word embeddings and embeddings from other modalities (e.g. image, audio) which are then fed to a BERT encoder. M-BERT has been recently adapted by \citet{jin2020complaint} for binary complaint prediction by inducing linguistic information instead of speech and image, however it did not perform better than BERT in their setting.

\renewcommand{\arraystretch}{1.1}
\begin{table*}[!t]
\begin{center}
\begin{tabular}{|l|c|c|c|c|}
\hline \rowcolor{dGray} \bf Model & \bf Acc & \bf P & \bf R & \bf F1\\ \hline

Majority Class & 35.2 & 8.8 & 25.0 & 13.0 \\
\hline

LR-BOW  & 46.7 $\pm$ .03 & 44.3 $\pm$ .06 & 43.6 $\pm$ .03 & 43.5 $\pm$ .03\\ 
BiGRU-Att  & 46.1 $\pm$ .03 & 43.6 $\pm$ .03 & 42.7 $\pm$ .02 & 43.5 $\pm$ .03\\  
RoBERTa & 58.7 $\pm$ .03 & 55.8 $\pm$ .05 & 55.4 $\pm$ .03 & 54.7 $\pm$ .04\\
\hline 
M-RoBERTa$_{Emo}$ & $\bf 59.8$ $\pm$ .02 & $\bf 56.6$ $\pm$ .03 & 55.7 $\pm$ .03 & $\bf 55.7^ \dagger $ $\pm$ .03\\
M-RoBERTa$_{Top}$ & 59.0 $\pm$ .03 & 55.9 $\pm$ .04 & 55.6 $\pm$ .03 & 55.2 $\pm$ .04\\
M-RoBERTa$_{Emo+Top}$ & 59.4 $\pm$ .03 & 56.5 $\pm$ .03 & $\bf 56.2$ $\pm$ .03 & 55.5 $\pm$ .02\\
\hline
\end{tabular}
\end{center}
\caption{\label{result1} Accuracy (Acc), Precision (P), Recall (R) and macro F1-Score (F1) for complaint severity level prediction ($\pm$ std. dev.). Best results are in bold. $\dagger$ indicates statistically significant improvement over RoBERTa (t-test, p<0.05).}
\end{table*}

We adapt M-BERT by replacing (1) the underlying BERT model with RoBERTa; and (2) the multimodal information with linguistic information. We first use a fully connected layer to project the linguistic representations into vectors with comparable size to RoBERTa's embeddings. Then we concatenate word representations from RoBERTa and linguistic information representations using a Multimodal Shifting Gate \citep{wang2019words}, where an attention gating mechanism is applied to control the influence of each representation. Finally, we apply layer normalization and dropout after the Multimodal Shifting Gate and pass the output to RoBERTa. We add an output layer to M-RoBERTa for classification similar to the RoBERTa model. We use M-RoBERTa with three types of linguistic features (i.e. emotion, topic and their combination):

\paragraph{M-RoBERTa$_{Emo}$}
We first use emotional information obtained by using a pretrained emotional classifier by \citet{volkova2016inferring}. This is 9-dimensional vector representing scores of sentiment (positive, negative and neutral) and six basic emotions of \citet{ekman1992argument} (anger, disgust, fear, joy, sadness and surprise). 

\paragraph{M-RoBERTa$_{Top}$}
We also use topical information from a 200-dimensional vector representing the distribution of the fraction of tokens in each tweet belonging to a topic cluster \citep{preoctiuc2015analysis,aletras2018predicting}.

\paragraph{M-RoBERTa$_{Emo+Top}$}
We finally experiment with injecting both emotional and topical information to M-RoBERTa.

\section{Experimental Setup}

\subsection{Hyperparameters} 

The {\bf BiGRU-Att} model uses 200-dimensional GloVe embeddings \citep{pennington2014glove} pre-trained on Twitter data. Its hidden size is $h$ = 128, $h \in$ \{64, 128, 256, 512\} with dropout $d$ = .2, $d \in$ \{.2, .5\}. We use Adam optimizer \citep{kingma2014adam} with learning rate $l$ = 1e-3, $l \in$ \{1e-3, 5e-3, 1e-2\}. For {\bf RoBERTa}, we use the Base uncased model and fine-tuning it with learning rate $l$ = 5e-6, $l \in$ \{1e-4, 1e-5, 5e-6, 1e-6\}. The maximum sequence length is set to 50 covering 95\% of tweets in the training set. For {\bf M-RoBERTa} models, we project the linguistic features (emotions and topics) to vectors of size $l$ = 200, $l \in$ \{200, 300, 400, 768\}. We also use dropout $d$ = .5, $d \in$ \{.2, .5\}. For all models we use a categorical-cross entropy loss following a similar approach to \citet{sun2019fine} which have achieved best results on fine-grained sentiment analysis (i.e. similar to the ordinal scale of complaints severity).

\subsection{Training and Evaluation} 
We run all models using a nested 10-fold cross validation approach, which consists of 2 nested loops as in \citet{preotiuc2019automatically}. In the outer loop, 9-folds are used for training and one for testing; while in the inner loop, a 3-fold cross validation is applied on the data from the nine folds (in the outer loop), where 2-folds are used for training and one for validation. During training, we choose the model with the smallest validation loss over 30 epochs. We measure predictive performance using the mean Accuracy, Precision, Recall and macro F1 over 10 folds (we also report the standard deviations).

\section{Results}

Table \ref{result1} shows the performance of all models including baselines and M-RoBERTa combined with linguistic information on complaint severity level prediction. 

Overall, M-RoBERTa models with linguistic features achieve best results. M-RoBERTa$_{Emo}$ outperforms all other models and reaches macro F1 up to 55.7. This confirms out hypothesis that injecting extra emotion information helps improve the performance of complaint severity level prediction. This is also in line with \citet{trosborg2011interlanguage} who states that the expression of complaints is relevant to different emotional states. The results of M-RoBERTa$_{Top}$ and M-RoBERTa$_{Emo+Top}$ are comparable with 55.2 and 55.5 macro F1 respectively. RoBERTa performs competitively but worse than the M-RoBERTa models. We also notice that BiGRU-Att does not perform well in our task (43.5 macro F1), which may result from the fact that it has not been pretrained.

\begin{figure}[!t]
\centering                                       
\includegraphics[scale=0.55]{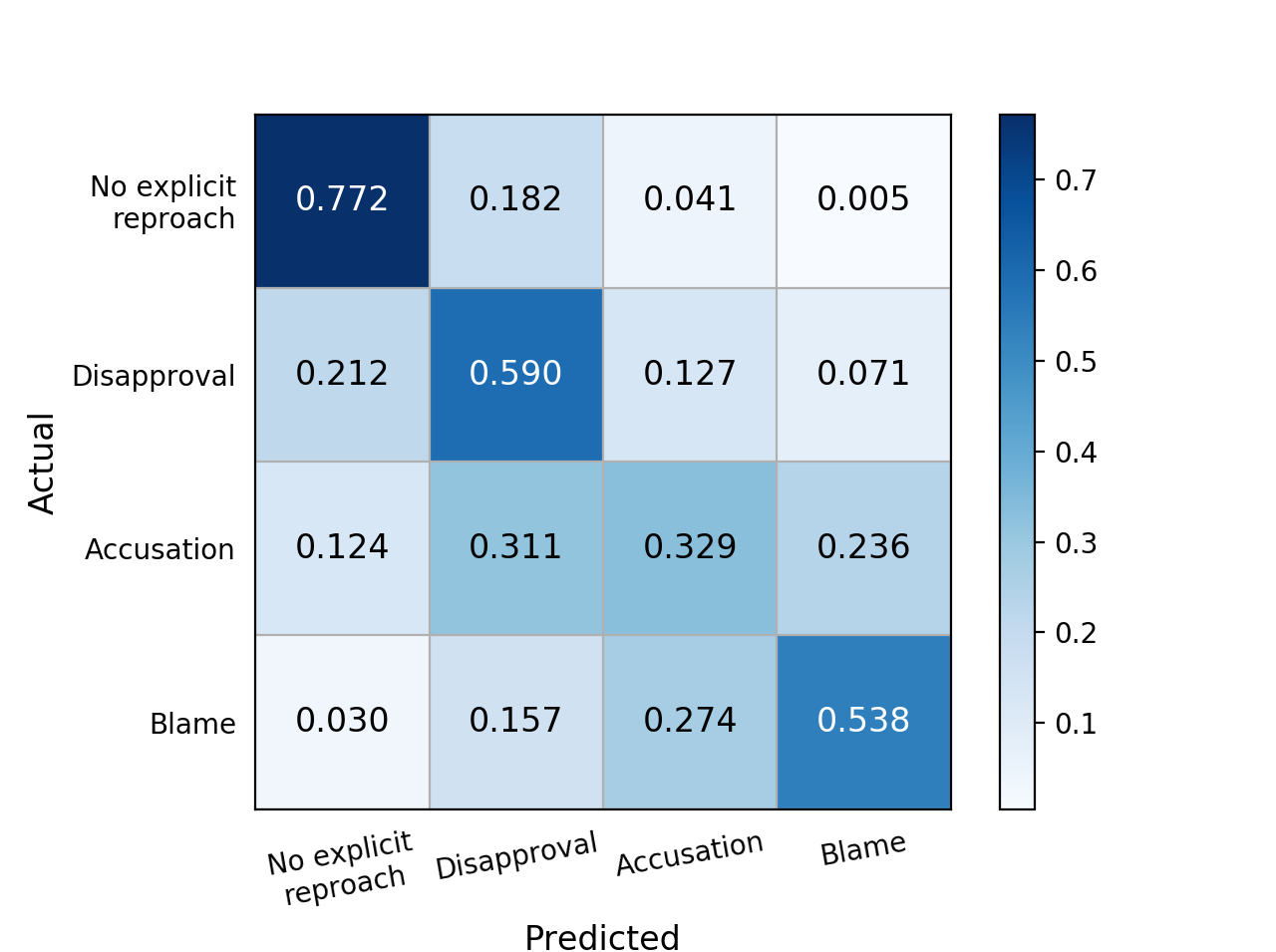}
\caption{Confusion matrix of the best performing model (M-RoBERTa$_{Emo}$).}             
\label{fig:roberta_cm} 
\end{figure}

Figure \ref{fig:roberta_cm} presents the confusion matrix of our best model (i.e. M-RoBERTa$_{Emo}$). The confusion matrix is normalized over the actual values (rows). The `No Explicit Reproach' category has the highest percentage (77.2\%) of correctly classified data points by the model, followed by label `Disapproval' with 59.0\%. These are also the two most frequent classes in the data set. On the other hand, results on `Accusation' are the lowest (32.9\%) which is confused with adjacent categories (`Disapproval' and `Blame'). Furthermore, the differences between mis-classifications and correct classification are relatively large for `Blame'. We speculate that this is because of the unique linguistic characteristic of the `Blame' category which gives emphasis on someone's responsibility. Finally, a category is more likely, in general, to be mis-classified to its adjacent severity categories. For example, when predicting `Disapproval', the number of model mis-classifications as `No Explicit Reproach' and `Accusation' is larger than `Blame'. This hints that tweets belonging to neighboring levels share more semantic, syntactic and stylistic similarities.

\begin{figure}[!t]
\center
\includegraphics[scale=0.55]{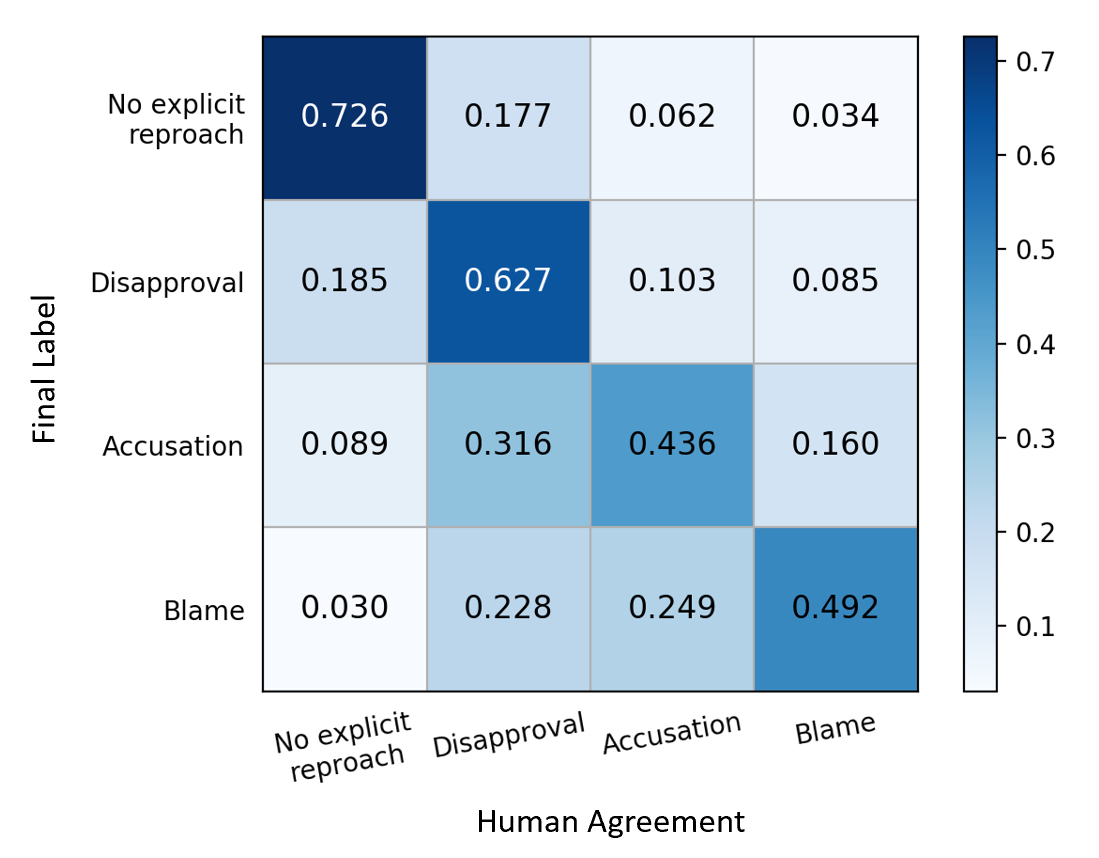}
\caption{Confusion matrix of human agreement.}             
\label{fig:annotator_cm} 
\end{figure}


We also compare the performance of our best model (i.e. M-RoBERTa$_{Emo}$) with human agreement for each class (Figure \ref{fig:annotator_cm}).  In general, the results of the model (shown in Figure \ref{fig:roberta_cm}) correlate to human agreement. In other words, the model and humans agree in the categories they confuse. For instance, it is easy for both of them to confuse `Accusation' with `Disapproval' (32.9\% vs. 31.1\% for the model and 43.6\% vs. 31.\% for humans). However, we observe that annotators are better at distinguishing high severity complaints from `No Explicit Reproach', where 21.2\% `Disapproval' and 12.4\% `Accusation' are wrongly classified as `No Explicit Reproach' by the model while the corresponding values are 18.5\% and 8.9\% by humans respectively. We argue that this is because annotators are able to identify subtle language (More details will be discussed in Section \ref{sec:analysis}). Also, we notice that the model achieves better performance when predicting `Blame', indicating a better capability on capturing the main characteristics of this class compared to humans.

\section{Error Analysis}
\label{sec:analysis}
We perform an error analysis to shed light on the limitations of our best performing model (M-RoBERTa$_{Emo}$) on complaint severity level classification. 

Firstly, we observe that most errors happen when the differences of tweets belonging to `Accusation' are blurred with `Disapproval' and `Blame'. The following two tweets are typical examples for `Accusation' being mis-classified as `Disapproval' and `Blame' respectively:
\begin{quote}
    \small
  \emph{$<$USER$>$, thank you ! Clear guidelines here, but {\bf not at all} what your advisor on the phone stated!}  
\end{quote}
\begin{quote}
    \small
  \emph{The new {\bf $<$USER$>$} stinks ...10mins to {\bf take my order} and another 15 to get it. And {\bf stop asking} my name like we're friends $<$URL$>$}  
\end{quote}
This is because some tweets belonging to `Accusation' also contain negation (e.g. `not at all') or negative terms (e.g. `disappointed'), which appear frequently in `Disapproval'. Also, consistent with the definition by \citet{trosborg2011interlanguage} (directly or indirectly accuses someone for causing the problem), tweets belonging to `Accusation' may involve doing something and contain terms like `$<$USER$>$' or `you', which is similar to complaints labeled as `Blame' such as:
\begin{quote}
    \small
  \emph{Thanks {\bf $<$USER$>$} for {\bf selling} expired beer \#fail {\bf $<$USER$>$} $<$URL$>$}  
\end{quote}

Secondly, the model struggles with complaints expressed in more subtle ways. In the following two examples, tweets belonging to `Disapproval' and `Accusation' are mis-classified as `No Explicit Reproach' respectively:
\begin{quote}
    \small
  \emph{Think someone at $<$USER$>$ had been drinking the stuff before they put the label on}  
\end{quote}
\begin{quote}
    \small
  \emph{Just opened a fresh bud light that was filled with water. Please explain $<$USER$>$.}  
\end{quote}
Such complaints do not contain terms that are typical of any specific complaint severity category (e.g. negation and negative terms in `Disapproval', person pronouns and terms describing undesirable results in `Blame') thus predicting them correctly needs more contextual understanding.

Finally, compared to other categories, the model is more likely to confuse tweets belonging to `No Explicit Reproach' and `Disapproval'. This happens because some tweets express weak dissatisfaction, which is difficult to identify. The following tweet is mis-classified as `No Explicit Reproach':
\begin{quote}
    \small
  \emph{Dearest $<$USER$>$: there really needs to be an easier method to report names that are inappropriate $<$URL$>$}  
\end{quote}
The model might need to learn more contextual information about such tweets instead of capturing certain relevant terms. Also, these two labels contain more similar terms such as `dm', `please help', `can't work' and interrogative tone. Examples of a `No Explicit Reproach' and `Disapproval' are the following (where similarities are in bold):
\begin{quote}
    \small
    \emph{Hey guys, I love this product featured on $<$USER$>$ today {\bf but} don't see a price? {\bf Help} a girl out{\bf ?} $<$URL$>$} 
\end{quote}  
\begin{quote}
    \small
    \emph{So it's going to cost \$7000 to fix the exhaust on my $<$USER$>$ 2009 jetta, and {\bf only} \$300 is covered under warranty. {\bf Help} $<$USER$>${\bf ?}}
\end{quote}

\section{Multi-task Learning for Binary Complaint Prediction}
We further experiment with multi-task learning (MTL) \citep{caruana1997multitask} for using severity categories to improve binary complaint prediction (i.e. complaint or non-complaint). MTL enables two or more tasks to be learned jointly by sharing information and parameters of a model. 

We explore whether or not the severity level of a complaint helps in complaint identification. We use the same data set as \citet{preotiuc2019automatically}, where each tweet is annotated as a complaint or not and our severity level annotations.\footnote{For a tweet that is a non-complaint, we assign an extra class for severity (i.e. `No Complaint Severity').}

\renewcommand{\arraystretch}{1.1}
\begin{table*}[!t]
\small
\begin{center}
\begin{tabular}{|l|c|c|c|c|}
\hline \rowcolor{dGray} \bf Model & \bf Acc & \bf P & \bf R & \bf F1\\ \hline
\rowcolor{mGray}  {\bf Single-task Learning} &&&& \\ 
LR-BOW+DS \citep{preotiuc2019automatically}  & 81.2 & - & -  & 79.0 \\
BiGRU-Att & 79.2 $\pm$ .05 & 79.2 $\pm$ .06 & 74.5 $\pm$ .05 & 74.5 $\pm$ .05\\
RoBERTa \citep{jin2020complaint} & 87.6 $\pm$ .03 & 86.6 $\pm$ .03 & 86.9 $\pm$ .03 & 86.6 $\pm$ .03\\ 
BERT \citep{jin2020complaint} & 88.0 $\pm$ .03 & 87.1 $\pm$ .03 & 87.3 $\pm$ .04 & 87.0 $\pm$ .03 \\
\rowcolor{mGray}  {\bf Multi-task Learning} &&&& \\ 
MTL-BiGRU-Att & 77.2 $\pm$ .05 & 75.4 $\pm$ .04 & 75.7 $\pm$ .04 & 75.4 $\pm$ .04\\
MTL-BiGRU-Att-DE & 75.7 $\pm$ .05 & 74.1 $\pm$ .05 & 74.6 $\pm$ .04 & 74.1 $\pm$ .05 \\
\hline
\citet{rajamanickam2020joint} &&&& \\
\hspace{.2cm} MTL-Hard Sharing & 75.2 $\pm$ .04 & 73.5 $\pm$ .05 & 71.5 $\pm$ .04 & 72.1 $\pm$ .05 \\
\hspace{.2cm} MTL-Double Encoder & 74.6 $\pm$ .03 & 72.7 $\pm$ .04 & 71.7 $\pm$ .03 & 72.0 $\pm$ .04 \\
\hspace{.2cm} MTL-Gated Double Encoder & 74.7 $\pm$ .03 & 73.4 $\pm$ .04 & 70.4 $\pm$ .03 & 71.1 $\pm$ .03 \\
\hline
\textbf{Ours} &&&& \\
\hspace{.2cm} MTL-M-RoBERTa$_{Emo}$ & $\bf 89.0$ $\pm$ .04 & 88.2 $\pm$ .03 & $\bf 88.4$ $\pm$ .03 & $\bf 88.2^ \dagger $ $\pm$ .03 \\
\hspace{.2cm} MTL-M-RoBERTa$_{Emo}$-DE & 88.9 $\pm$ .04 & $\bf 88.3$ $\pm$ .04 & 88.3 $\pm$ .03 & 88.1 $\pm$ .04 \\
\hline
\end{tabular}
\end{center}
\caption{\label{result2} Accuracy (Acc), Precision (P), Recall (R) and macro F1-Score (F1) for binary complaint prediction ($\pm$ std. dev.). Best results are in bold. $\dagger$ indicates statistically significant improvement over BERT \citep{jin2020complaint} in STL (t-test, p<0.05).}
\end{table*}

\subsection{Predictive models}
We first adapt three multi-task learning models based on bidirectional recurrent neural networks recently proposed by \citet{rajamanickam2020joint} for jointly modeling abusive language detection and emotion detection. We also adapt our {M-RoBERTa$_{Emo}$} model in a multi-task setting using two variants. We use the severity complaint prediction as an auxiliary task and the binary complaint prediction as the main task to train different MTL models. All models are trained on the two tasks and updated at the same time with a joint loss: $$L=(1-\alpha)L_{com}+\alpha L_{sev}$$
\noindent where $L_{com}$ and $L_{sev}$ are the losses of complaint identification and severity level classification tasks respectively. $\alpha$ is a parameter to control the importance of each loss.

\paragraph{MTL-Hard Sharing} We adapt the MTL-Hard Sharing model of \citet{rajamanickam2020joint}, where a single encoder is shared and updated by both tasks. We first pass GloVe embedding representations to a shared stacked BiGRU encoder. Then the output of the shared encoder is fed to two different BiGRU-Att models specific to each task (complaint detection and severity level identification) separately. Finally, we add an output layer with a sigmoid and a softmax activation function for binary and multi-class prediction respectively.

\paragraph{MTL-Double Encoder} Instead of sharing a single encoder, the MTL-Double Encoder model \citep{rajamanickam2020joint} utilizes two stacked BiGRU encoders, where one is task-specific (complaint detection only) and the another one is shared by both tasks. We pass the output of the shared encoder to a BiGRU-Att model for severity level prediction. We also concatenate the output of the task-specific and shared encoder and pass it to another BiGRU-Att model for complaint prediction. The rest of the architecture is the same as the MTL-Hard Sharing model.

\paragraph{MTL-Gated Double Encoder} The MTL-Gated Double Encoder model \citep{rajamanickam2020joint} has the same architecture as the MTL-Double Encoder. The outputs from two stacked BiBRU-Att encoders are concatenated by assigning a weight to each representation [($1-\beta$) for the output of the task-specific encoder layer and $\beta$ for the output of the shared one)] that controls the importance of the two representations.

\paragraph{MTL-M-RoBERTa$_{Emo}$} We also adapt our best performing model in the severity prediction task (M-RoBERTa$_{Emo}$) to support multi-task learning by adding an extra output layer for binary complaint prediction (MTL-M-RoBERTa$_{Emo}$). 

\paragraph{MTL-M-RoBERTa$_{Emo}$-DE} We pass the M-RoBERTa$_{Emo}$ embedding to two separate RoBERTa encoders, i.e. double encoder (DE), followed by two classifiers for binary complaint and severity level prediction (MTL-M-RoBERTa$_{Emo}$-DE). 

\subsection{Experimental Setup}

\paragraph{Baselines} We compare MTL models with the following baselines on binary complaint identification: (1) Logistic Regression with bag-of-words using distant supervision\footnote{Distant supervision method uses larger `noisy' data to further boost the performance of the model} ({\bf LR-BOW + DS}) \citep{preotiuc2019automatically}; (2) A  standard BiGRU-Att model ({\bf BiGRU-Att}); (3) A RoBERTa base model without combining linguistic information ({\bf RoBERTa}) \citep{jin2020complaint}; (4) A BERT base model without combining linguistic information ({\bf BERT}) which has been shown to achieve state-of-the-art results in binary complaint identification \citep{jin2020complaint}; and (5) replacing M-RoBERTa$_{Emo}$ with BiGRU-Att in the MTL-M-RoBERTa$_{Emo}$ and MTL-M-RoBERTa$_{Emo}$-DE models ({\bf MTL-BiGRU-Att} and {\bf MTL-BiGRU-Att-DE}).

\paragraph{Hyperparameters}
We train\footnote{We experiment with all MTL models using the same training and evaluation method as in the complaint severity prediction task.} the {\bf MTL-BiGRU-Att} and {\bf MTL-BiGRU-Att-DE} model with the same hyperparameters as BiGRU-Att in complaint severity prediction. For the {\bf MTL-Hard Sharing}, {\bf MTL-Double Encoder} and {\bf MTL-Gated Double Encoder} model, the hidden size of the stacked BiGRU encoder(s) and BiGRU-Att models is $h$ = 128, $h \in$ \{64, 128, 256, 512\}. We set $\beta$ in {\bf MTL-Gated Double Encoder} and the remaining parameters in three models to be the same as \citet{rajamanickam2020joint}. We train {\bf MTL-M-RoBERTa$_{Emo}$} and {\bf MTL-M-RoBERTa$_{Emo}$-DE} with a learning rate $l$ = 1e-6, $l \in$ \{1e-5, 5e-6, 1e-6\}. The rest of the parameters is the same as M-RoBERTa$_{Emo}$ in the complaint severity prediction. The parameter $\alpha$ which controls the importance of the two losses is set to .1, $\alpha \in$ \{.001, .01, .1, .3, .5\}.

\subsection{Results}
Table \ref{result2} shows results of the single-task learning (STL) and multi-task learning (MTL) models on the complaint identification task. Overall, we observe that all MTL models using M-RoBERTa$_{Emo}$ perform better than the majority of STL models, indicating severity detection improves binary complaint identification. MTL-M-RoBERTa$_{Emo}$ outperforms all other models achieving 88.2 macro F1, followed by MTL-M-RoBERTa$_{Emo}$-DE with 88.1 F1. This confirms our hypothesis that complaint identification can be benefited by the complaint severity level information when jointly learning these two tasks simultaneously. Also, MTL-BiGRU-Att performs better than BiGRU-Att in STL achieving 75.4 F1 while the results of BiGRU-Att (74.5 F1) and MTL-BiGRU-Att-DE (74.1 F1) are comparable. We notice that the models proposed by \citet{rajamanickam2020joint} (i.e. MTL-Hard sharing, MTL-Double Encoder and MTL-Gated Double Encoder) achieve low performance with only the MTL-Hard Sharing model performing slightly better than the others with 72.1 macro F1. We speculate that adding one or more extra BiGRU encoders before the BiGRU-Att model is an overly complex structure for our data set.

\renewcommand{\arraystretch}{1.2}
\begin{table*}[!t]
\small
\center
\resizebox{\textwidth}{!}{
\begin{tabular}{|p{5cm}|c|c|c|c|}
\hline
\rowcolor{dGray} \bf Tweet & \bf BERT & \bf MTL-M-RoBERTa{$_{Emo}$} & Actual Label & Severity Label \\
\hline
\emph{What's your secret to poaching eggs? Mine {\bf never} look that good.} & Complaint & Non-complaint & Non-complaint &  No Complaint Severity \\
\hline
\rowcolor{lGray} \emph{$<$URL$>$ How {\bf bad} do you really want a ps4 this year? Get a pre-owned playstation 4 at a {\bf very low} dis $<$URL$>$} & Complaint & Non-complaint & Non-complaint &  No Complaint Severity \\
\hline
\emph{So, I'm now having to check my $<$USER$>$ forester's oil each month. Put 4 quarts in today, got about 2 out. \#smh} & Non-complaint & Complaint & Complaint & Disapproval \\
\hline
\rowcolor{lGray} \emph{ls this how you fix the exhaust of your $<$USER$>$ in \#belarus? $<$URL$>$} & Non-complaint & Complaint & Complaint & Blame\\
\hline
\end{tabular}}
\caption{Complaint classification examples by BERT \citep{jin2020complaint} and our MTL-M-RoBERTa{$_{Emo}$} compared to the actual labels.}
\label{t:ExampleMTL}
\end{table*}

\subsection{Analysis}
We investigate the influence of recognizing severity levels of complaints on binary complaint identification in our MTL setting. We analyze predictive results by inspecting predictions from the previous best performing model BERT (STL) and MTL-M-RoBERTa$_{Emo}$ models in a random fold (out of 10 CV folds). We observe that 9.8\% of predictions flip, where the number of complaints flipping to non-complaints is noticeably larger (88.2\%) than that of non-complaints flipping to complaints (11.8\%). Similarly, we also compare predicted results between BiGRU-Att (STL) and MTL-BiGRU-Att in the same fold. The flipping percentage (6.9\%) is lower than BERT and MTL-M-RoBERTa$_{Emo}$ while the proportions of one class flipping to another are consistent (83.4\% and 16.6\% respectively). These indicate that complaint severity information encapsulates complementary information for the model to predict non-complaints accurately. 

Table \ref{t:ExampleMTL} shows flipping examples from BERT (STL) and MTL-M-RoBERTa$_{Emo}$. From the first two rows, we see that the MTL model is not affected by negation (e.g. `never') and negative terms (e.g. `bad', `very low') using the extra knowledge provided by the severity level prediction task. Also, in the last two examples, complaints are expressed in a more subtle way that rarely contains typical complaint-related terms. This indicates the MTL model is able to detect this type of complaints correctly because the severity level information encourages the model to learn to distinguish between such stylistic idiosyncrasies.

We further observe that 11.2\% of wrong predictions remain the same for the two models, where complaints and non-complaints account for 59.0\% and 41.0\% respectively which means severity features benefit more posts that are complaints to be classified accurately. On the other hand, the model still has difficulty in predicting some non-complaint posts which might happen because of the lower performance of severity detection\footnote{Severity prediction is less accurate in MTL than in a single task setting.} when used as an auxiliary task in the MTL setting.

\section{Conclusion}
We presented the first study on severity level of complaints in computational linguistics. We developed a publicly available data set of tweets labeled with four categories based on theory of pragmatics. We modeled complaint severity level prediction as a new multi-class classification task and conducted experiments using different transformer-based networks combined with linguistic features reaching up to 55.7 macro F1. We further used a multi-task learning setting to jointly model binary complaint prediction and complaint severity classification as an auxiliary task achieving new state-of-the-art performance on complaint detection (88.2 macro F1). 
In the future, we plan to apply our methods on a multilingual setting across different platforms. 

\subsection*{Acknowledgments}
We would like to thank Areej Alokaili, Daniel Preo\c{t}iuc-Pietro and Danae Sanchez Villegas for their helpful feedback. NA is supported by ESRC grant ES/T012714/1.

\subsection*{Ethics Statement}
Our work has received approval from the Ethics Committee of our institution (Ref. No 031504) and complies with Twitter data policy for research.\footnote{\url{https://developer.twitter.com/en/developer-terms/agreement-and-policy}} 

\bibliography{anthology,custom}

\begin{thebibliography}{43}
\expandafter\ifx\csname natexlab\endcsname\relax\def\natexlab#1{#1}\fi

\bibitem[{Akhtar et~al.(2020)Akhtar, Ekbal, and Cambria}]{akhtar2020intense}
Md~Shad Akhtar, Asif Ekbal, and Erik Cambria. 2020.
\newblock \href {https://doi.org/10.1109/MCI.2019.2954667} {How intense are
  you? {P}redicting intensities of emotions and sentiments using stacked
  ensemble}.
\newblock \emph{IEEE Computational Intelligence Magazine}, 15(1):64--75.

\bibitem[{Alejo et~al.(2020)Alejo, Badia, and Barnes}]{alejo2020cross}
Irean~Navas Alejo, Toni Badia, and Jeremy Barnes. 2020.
\newblock Cross-lingual {E}motion {I}ntensity {P}rediction.
\newblock \emph{arXiv preprint arXiv:2004.04103}.

\bibitem[{Aletras and Chamberlain(2018)}]{aletras2018predicting}
Nikolaos Aletras and Benjamin~Paul Chamberlain. 2018.
\newblock Predicting twitter user socioeconomic attributes with network and
  language information.
\newblock In \emph{Proceedings of the 29th ACM International Conference on
  Hypertext and Social Media}, pages 20--24.

\bibitem[{Artstein and Poesio(2008)}]{artstein2008inter}
Ron Artstein and Massimo Poesio. 2008.
\newblock \href {https://doi.org/10.1162/coli.07-034-R2} {Inter-{C}oder
  {A}greement for {C}omputational {L}inguistics}.
\newblock \emph{Computational Linguistics}, 34(4):555--596.

\bibitem[{Bhat and Culotta(2017)}]{bhat2017identifying}
Shreesh~Kumara Bhat and Aron Culotta. 2017.
\newblock Identifying leading indicators of product recalls from online reviews
  using positive unlabeled learning and domain adaptation.
\newblock In \emph{Eleventh International AAAI Conference on Web and Social
  Media}.

\bibitem[{Caruana(1997)}]{caruana1997multitask}
Rich Caruana. 1997.
\newblock \href {https://doi.org/10.1023/A:1007379606734} {Multitask
  {L}earning}.
\newblock \emph{Machine learning}, 28(1):41--75.

\bibitem[{Cho et~al.(2014)Cho, Van~Merri{\"e}nboer, Gulcehre, Bahdanau,
  Bougares, Schwenk, and Bengio}]{cho2014learning}
Kyunghyun Cho, Bart Van~Merri{\"e}nboer, Caglar Gulcehre, Dzmitry Bahdanau,
  Fethi Bougares, Holger Schwenk, and Yoshua Bengio. 2014.
\newblock Learning {P}hrase {R}epresentations using {RNN} {E}ncoder-{D}ecoder
  for {S}tatistical {M}achine {T}ranslation.
\newblock \emph{arXiv preprint arXiv:1406.1078}.

\bibitem[{Coussement and Van~den Poel(2008)}]{coussement2008improving}
Kristof Coussement and Dirk Van~den Poel. 2008.
\newblock \href {https://doi.org/10.1016/j.dss.2007.10.010} {Improving
  {C}ustomer {C}omplaint {M}anagement by {A}utomatic {E}mail {C}lassification
  {U}sing {L}inguistic {S}tyle {F}eatures as {P}redictors}.
\newblock \emph{Decision Support Systems}, 44(4):870--882.

\bibitem[{Danisman and Alpkocak(2008)}]{danisman2008feeler}
Taner Danisman and Adil Alpkocak. 2008.
\newblock Feeler: Emotion {C}lassification of {T}ext {U}sing {V}ector {S}pace
  {M}odel.
\newblock In \emph{AISB 2008 Convention Communication, Interaction and Social
  Intelligence}, volume~1, page~53.

\bibitem[{Devlin et~al.(2018)Devlin, Chang, Lee, and
  Toutanova}]{devlin2018bert}
Jacob Devlin, Ming-Wei Chang, Kenton Lee, and Kristina Toutanova. 2018.
\newblock Bert: Pre-training of {D}eep {B}idirectional {T}ransformers for
  {L}anguage {U}nderstanding.
\newblock In \emph{Proceedings of the 2019 Conference of the North American
  Chapter of the Association for Computational Linguistics: Human Language
  Technologies, Volume 1 (Long and Short Papers)}, page 4171–4186.

\bibitem[{Ekman(1992)}]{ekman1992argument}
Paul Ekman. 1992.
\newblock \href {https://doi.org/10.1080/02699939208411068} {An {A}rgument for
  {B}asic {E}motions}.
\newblock \emph{Cognition \& Emotion}, 6(3-4):169--200.

\bibitem[{Fleiss(1971)}]{fleiss1971measuring}
Joseph~L Fleiss. 1971.
\newblock Measuring nominal scale agreement among many raters.
\newblock \emph{Psychological bulletin}, 76(5):378.

\bibitem[{Forster and Entrup(2017)}]{forster2017cognitive}
J~Forster and B~Entrup. 2017.
\newblock \href {https://doi.org/10.1088/1757-899X/261/1/012016} {A {C}ognitive
  {C}omputing {A}pproach for {C}lassification of {C}omplaints in the
  {I}nsurance {I}ndustry}.
\newblock In \emph{IOP Conference Series: Materials Science and Engineering},
  volume 261, page 012016. IOP Publishing.

\bibitem[{Gunawan et~al.(2018)Gunawan, Siregar, Rahmat, and
  Amalia}]{gunawan2018building}
D~Gunawan, RP~Siregar, RF~Rahmat, and A~Amalia. 2018.
\newblock \href {https://doi.org/10.1088/1742-6596/978/1/012119} {Building
  {A}utomatic {C}ustomer {C}omplaints {F}iltering {A}pplication based on
  {T}witter in {B}ahasa {I}ndonesia}.
\newblock In \emph{Journal of Physics: Conference Series}, volume 978, page
  012119. IOP Publishing.

\bibitem[{Jin et~al.(2013)Jin, Yan, Yu, and Li}]{jin2013service}
Jiahua Jin, Xiangbin Yan, You Yu, and Yijun Li. 2013.
\newblock Service {F}ailure {C}omplaints {I}dentification in {S}ocial {M}edia:
  {A} {T}ext {C}lassification {A}pproach.

\bibitem[{Jin and Aletras(2020)}]{jin2020complaint}
Mali Jin and Nikolaos Aletras. 2020.
\newblock Complaint identification in social media with transformer networks.
\newblock In \emph{Proceedings of the 28th International Conference on
  Computational Linguistics}, pages 1765--1771.

\bibitem[{Kakolaki and Shahrokhi(2016)}]{kakolakigender}
Leila~Nasiri Kakolaki and Mohsen Shahrokhi. 2016.
\newblock Gender {D}ifferences in {C}omplaint {S}trategies among {I}ranian
  {U}pper {I}ntermediate {EFL} {S}tudents.

\bibitem[{Kingma and Ba(2014)}]{kingma2014adam}
Diederik~P Kingma and Jimmy Ba. 2014.
\newblock Adam: A {M}ethod for {S}tochastic {O}ptimization.
\newblock \emph{arXiv preprint arXiv:1412.6980}.

\bibitem[{Laksana and Purwarianti(2014)}]{laksana2014indonesian}
Janice Laksana and Ayu Purwarianti. 2014.
\newblock \href {https://doi.org/10.1109/ICAICTA.2014.7005928} {Indonesian
  {T}witter {T}ext {A}uthority {C}lassification for {G}overnment in {B}andung}.
\newblock In \emph{2014 International Conference of Advanced Informatics:
  Concept, Theory and Application (ICAICTA)}, pages 129--134. IEEE.

\bibitem[{Liu et~al.(2019)Liu, Ott, Goyal, Du, Joshi, Chen, Levy, Lewis,
  Zettlemoyer, and Stoyanov}]{liu2019roberta}
Yinhan Liu, Myle Ott, Naman Goyal, Jingfei Du, Mandar Joshi, Danqi Chen, Omer
  Levy, Mike Lewis, Luke Zettlemoyer, and Veselin Stoyanov. 2019.
\newblock {RoBERTa}: A {R}obustly {O}ptimized {BERT} {P}retraining {A}pproach.
\newblock \emph{arXiv preprint arXiv:1907.11692}.

\bibitem[{Maronikolakis et~al.(2020)Maronikolakis, Villegas,
  Preo{\c{t}}iuc-Pietro, and Aletras}]{maronikolakis2020analyzing}
Antonios Maronikolakis, Danae~S{\'a}nchez Villegas, Daniel
  Preo{\c{t}}iuc-Pietro, and Nikolaos Aletras. 2020.
\newblock Analyzing political parody in social media.
\newblock In \emph{Proceedings of the 58th Annual Meeting of the Association
  for Computational Linguistics}, pages 4373--4384.

\bibitem[{Merson and Mary(2017)}]{merson2017text}
Febina Merson and Roseline Mary. 2017.
\newblock A {T}ext {M}ining {A}pproach to {I}dentify and {A}nalyse {P}rominent
  {I}ssues from {P}ublic {C}omplaints.
\newblock \emph{International Journal of Advanced Research in Computer and
  Communication Engineering}, 6(3).

\bibitem[{Olshtain and Weinbach(1985)}]{olshtain1985complaints}
E~Olshtain and L~Weinbach. 1985.
\newblock Complaints: {A} {S}tudy of {S}peech {A}ct {B}ehavior among {N}ative
  and {N}onnative {S}peakers of {H}ebrew. {T}he {P}rag-matic {P}erspective.

\bibitem[{Pennington et~al.(2014)Pennington, Socher, and
  Manning}]{pennington2014glove}
Jeffrey Pennington, Richard Socher, and Christopher~D Manning. 2014.
\newblock Glo{V}e: {G}lobal {V}ectors for {W}ord {R}epresentation.
\newblock In \emph{Proceedings of the 2014 conference on empirical methods in
  natural language processing (EMNLP)}, pages 1532--1543.

\bibitem[{Preotiuc-Pietro et~al.(2019)Preotiuc-Pietro, Gaman, and
  Aletras}]{preotiuc2019automatically}
Daniel Preotiuc-Pietro, Mihaela Gaman, and Nikolaos Aletras. 2019.
\newblock \href {https://www.aclweb.org/anthology/P19-1495} {Automatically
  {I}dentifying {C}omplaints in {S}ocial {M}edia}.
\newblock In \emph{Proceedings of the 57th Annual Meeting of the Association
  for Computational Linguistics}, pages 5008--5019.

\bibitem[{Preo{\c{t}}iuc-Pietro et~al.(2015)Preo{\c{t}}iuc-Pietro, Lampos, and
  Aletras}]{preoctiuc2015analysis}
Daniel Preo{\c{t}}iuc-Pietro, Vasileios Lampos, and Nikolaos Aletras. 2015.
\newblock An analysis of the user occupational class through {T}witter content.
\newblock In \emph{Proceedings of the 53rd Annual Meeting of the Association
  for Computational Linguistics and the 7th International Joint Conference on
  Natural Language Processing (Volume 1: Long Papers)}, pages 1754--1764.

\bibitem[{Rahman et~al.(2019)Rahman, Hasan, Zadeh, Morency, and
  Hoque}]{rahman2019m}
Wasifur Rahman, Md~Kamrul Hasan, Amir Zadeh, Louis-Philippe Morency, and
  Mohammed~Ehsan Hoque. 2019.
\newblock {M-BERT}: {I}njecting {M}ultimodal {I}nformation in the {BERT}
  {S}tructure.
\newblock \emph{arXiv preprint arXiv:1908.05787}.

\bibitem[{Rajamanickam et~al.(2020)Rajamanickam, Mishra, Yannakoudakis, and
  Shutova}]{rajamanickam2020joint}
Santhosh Rajamanickam, Pushkar Mishra, Helen Yannakoudakis, and Ekaterina
  Shutova. 2020.
\newblock Joint {M}odelling of {E}motion and {A}busive {L}anguage {D}etection.
\newblock \emph{arXiv preprint arXiv:2005.14028}.

\bibitem[{Schwartz et~al.(2017)Schwartz, Giorgi, Sap, Crutchley, Ungar, and
  Eichstaedt}]{schwartz2017dlatk}
H~Andrew Schwartz, Salvatore Giorgi, Maarten Sap, Patrick Crutchley, Lyle
  Ungar, and Johannes Eichstaedt. 2017.
\newblock Dlatk: Differential {L}anguage {A}nalysis {T}oolkit.
\newblock In \emph{Proceedings of the 2017 Conference on Empirical Methods in
  Natural Language Processing: System Demonstrations}, pages 55--60.

\bibitem[{Sun et~al.(2019)Sun, Qiu, Xu, and Huang}]{sun2019fine}
Chi Sun, Xipeng Qiu, Yige Xu, and Xuanjing Huang. 2019.
\newblock How to {F}ine-{T}une {BERT} for {T}ext {C}lassification?
\newblock In \emph{China National Conference on Chinese Computational
  Linguistics}, pages 194--206. Springer.

\bibitem[{Tatsuki(2000)}]{tatsuki2000if}
Donna~Hurst Tatsuki. 2000.
\newblock \href {https://doi.org/10.1016/S0378-2166(99)00076-4} {If my
  complaints could passions move: An interlanguage study of aggression}.
\newblock \emph{Journal of Pragmatics}, 32(7):1003--1017.

\bibitem[{Tian et~al.(2018)Tian, Rong, Shi, Liu, and Xiong}]{tian2018attention}
Zhengxi Tian, Wenge Rong, Libin Shi, Jingshuang Liu, and Zhang Xiong. 2018.
\newblock Attention {A}ware {B}idirectional {G}ated {R}ecurrent {U}nit {B}ased
  {F}ramework for {S}entiment {A}nalysis.
\newblock In \emph{International Conference on Knowledge Science, Engineering
  and Management}, pages 67--78. Springer.

\bibitem[{Tjandra et~al.(2015)Tjandra, Warsito, and
  Sugiono}]{tjandra2015determining}
Suhatati Tjandra, Amelia Alexandra~Putri Warsito, and Judi~Prajetno Sugiono.
  2015.
\newblock \href {https://doi.org/10.1109/ICTKE.2015.7368461} {Determining
  {C}itizen {C}omplaints to the {A}ppropriate {G}overnment {D}epartments
  {U}sing {KNN} {A}lgorithm}.
\newblock In \emph{2015 13th International Conference on ICT and Knowledge
  Engineering (ICT \& Knowledge Engineering 2015)}, pages 1--4. IEEE.

\bibitem[{Trosborg(2011)}]{trosborg2011interlanguage}
Anna Trosborg. 2011.
\newblock \emph{Interlanguage pragmatics: Requests, complaints, and apologies},
  volume~7.
\newblock Walter de Gruyter.

\bibitem[{Tsakalidis et~al.(2018)Tsakalidis, Aletras, Cristea, and
  Liakata}]{tsakalidis2018nowcasting}
Adam Tsakalidis, Nikolaos Aletras, Alexandra~I Cristea, and Maria Liakata.
  2018.
\newblock Nowcasting the stance of social media users in a sudden vote: The
  case of the greek referendum.
\newblock In \emph{Proceedings of the 27th ACM International Conference on
  Information and Knowledge Management}, pages 367--376.

\bibitem[{Van~Noort and Willemsen(2012)}]{van2012online}
Guda Van~Noort and Lotte~M Willemsen. 2012.
\newblock \href {https://doi.org/10.1016/j.intmar.2011.07.001} {Online {D}amage
  {C}ontrol: The {E}ffects of {P}roactive {V}ersus {R}eactive {W}ebcare
  {I}nterventions in {C}onsumer-generated and {B}rand-generated {P}latforms}.
\newblock \emph{Journal of Interactive Marketing}, 26(3):131--140.

\bibitem[{V{\'a}squez(2011)}]{vasquez2011complaints}
Camilla V{\'a}squez. 2011.
\newblock \href {https://doi.org/10.1016/j.pragma.2010.11.007} {Complaints
  {O}nline: {T}he {C}ase of {T}ripadvisor}.
\newblock \emph{Journal of Pragmatics}, 43(6):1707--1717.

\bibitem[{Vaswani et~al.(2017)Vaswani, Shazeer, Parmar, Uszkoreit, Jones,
  Gomez, Kaiser, and Polosukhin}]{vaswani2017attention}
Ashish Vaswani, Noam Shazeer, Niki Parmar, Jakob Uszkoreit, Llion Jones,
  Aidan~N Gomez, {\L}ukasz Kaiser, and Illia Polosukhin. 2017.
\newblock Attention {I}s {A}ll {Y}ou {N}eed.
\newblock In \emph{Advances in neural information processing systems}, pages
  5998--6008.

\bibitem[{Volkova and Bachrach(2016)}]{volkova2016inferring}
Svitlana Volkova and Yoram Bachrach. 2016.
\newblock Inferring {P}erceived {D}emographics from {U}ser {E}motional {T}one
  and {U}ser-{E}nvironment {E}motional {C}ontrast.
\newblock In \emph{Proceedings of the 54th Annual Meeting of the Association
  for Computational Linguistics (Volume 1: Long Papers)}, pages 1567--1578.

\bibitem[{Wang et~al.(2019)Wang, Shen, Liu, Liang, Zadeh, and
  Morency}]{wang2019words}
Yansen Wang, Ying Shen, Zhun Liu, Paul~Pu Liang, Amir Zadeh, and Louis-Philippe
  Morency. 2019.
\newblock Words {C}an {S}hift: {D}ynamically {A}djusting {W}ord
  {R}epresentations {U}sing {N}onverbal {B}ehaviors.
\newblock In \emph{Proceedings of the AAAI Conference on Artificial
  Intelligence}, volume~33, pages 7216--7223.

\bibitem[{Xu et~al.(2017)Xu, Liu, Guo, Sinha, and Akkiraju}]{Xu2017new}
Anbang Xu, Zhe Liu, Yufan Guo, Vibha Sinha, and Rama Akkiraju. 2017.
\newblock \href {https://doi.org/10.1145/3025453.3025496} {A {N}ew {C}hatbot
  for {C}ustomer {S}ervice on {S}ocial {M}edia}.
\newblock In \emph{Proceedings of the Conference on Human Factors in Computing
  Systems (CHI)}, pages 3506--3510.

\bibitem[{Yang et~al.(2019)Yang, Tan, Lu, Cui, Li, Chen, Xiong, Wang, Li, Pei
  et~al.}]{yang2019detecting}
Wei Yang, Luchen Tan, Chunwei Lu, Anqi Cui, Han Li, Xi~Chen, Kun Xiong, Muzi
  Wang, Ming Li, Jian Pei, et~al. 2019.
\newblock Detecting {C}ustomer {C}omplaint {E}scalation with {R}ecurrent
  {N}eural {N}etworks and {M}anually-{E}ngineered {F}eatures.
\newblock In \emph{Proceedings of the 2019 Conference of the North American
  Chapter of the Association for Computational Linguistics: Human Language
  Technologies, Volume 2 (Industry Papers)}, pages 56--63.

\bibitem[{Zhang et~al.(2018)Zhang, Fu, She, Zhang, Wang, and
  Yang}]{zhang2018ijcai}
Yuxiang Zhang, Jiamei Fu, Dongyu She, Ying Zhang, Senzhang Wang, and Jufeng
  Yang. 2018.
\newblock \href {https://doi.org/10.24963/ijcai.2018/639} {Text {E}motion
  {D}istribution {L}earning via {M}ulti-{T}ask {C}onvolutional {N}eural
  {N}etwork}.
\newblock In \emph{Proceedings of the Twenty-Seventh International Joint
  Conference on Artificial Intelligence, {IJCAI-18}}, pages 4595--4601.

\end{thebibliography}
\bibliographystyle{acl_natbib}







\end{document}